\pdfoutput=1

\documentclass[11pt]{article}

\usepackage[]{acl}

\usepackage{times}
\usepackage{latexsym}

\usepackage[T1]{fontenc}

\usepackage[utf8]{inputenc}

\usepackage{microtype}

\usepackage{amsmath,amsfonts,amssymb}
\usepackage{multirow}
\usepackage{xspace}
\usepackage{subcaption}
\usepackage{esint}
\usepackage{appendix}
\usepackage{siunitx}
\usepackage{etoolbox}
\usepackage{stfloats}
\usepackage{graphicx}
\usepackage[ruled,vlined]{algorithm2e}
\usepackage{enumitem}
\usepackage{algpseudocode}
\usepackage{floatrow}
\usepackage{bm}
\usepackage{bbm}
\newfloatcommand{capbtabbox}{table}[][\FBwidth]

\usepackage{blindtext}
\usepackage{enumitem}
\usepackage{wrapfig}
\usepackage{booktabs}
\usepackage{xspace}
\usepackage{placeins}
\usepackage{amsmath}
\usepackage{amssymb}
\usepackage{mathtools}
\usepackage{amsthm}
\usepackage{amsfonts}
\usepackage[capitalize,noabbrev]{cleveref}
\usepackage{caption}

\theoremstyle{plain}

\theoremstyle{definition}

\theoremstyle{remark}

\newcommand{\system}[1]{\textsc{#1}\xspace}
\newcommand{\data}[1]{\textsc{#1}\xspace}

\newcommand{\btmetric}[1]{BT Discrepancy Rate}

\newcommand{\geode}{\data{GeoQuery(De)}}
\newcommand{\geoel}{\data{GeoQuery(El)}}
\newcommand{\geoth}{\data{GeoQuery(Th)}}

\newcommand{\geo}{\data{GeoQuery}}
\newcommand{\nlmap}{\data{NLMap}}
\newcommand{\nlmapde}{\data{NLMap(De)}}

\newcommand{\bertlstm}{\system{BERT-LSTM}}

\newcommand{\rand}{\system{Random}}
\newcommand{\cluster}{\system{Cluster}}
\newcommand{\csse}{\system{CSSE}}
\newcommand{\rttl}{\system{RTTL}}
\newcommand{\lcsfw}{\system{LCS(FW)}}
\newcommand{\lcsbw}{\system{LCS(BW)}}
\newcommand{\traffic}{\system{Traffic}}
\newcommand{\oracle}{\system{Oracle}}
\newcommand{\amspnbest}{\system{ABE(n-best)}}
\newcommand{\amspmax}{\system{ABE(max)}}
\newcommand{\amsp}{\system{ABE}}
\newcommand{\hatt}{\system{HAT}}

\newcommand{\lfslcd}{\system{LFS-LC-D}}

\usepackage{amsthm,amsmath,amsfonts,bm,xspace}
\usepackage{color}

\newcommand{\comment}[1]{}

\def\eqref#1{(\ref{#1})}

\def\1{\bm{1}}

\def\vx{{\bm{x}}}
\def\vy{{\bm{y}}}

\DeclareMathAlphabet{\mathsfit}{\encodingdefault}{\sfdefault}{m}{sl}
\SetMathAlphabet{\mathsfit}{bold}{\encodingdefault}{\sfdefault}{bx}{n}

\newcommand{\KL}{D_{\mathrm{KL}}}

\DeclareMathOperator*{\argmax}{arg\,max}
\DeclareMathOperator*{\argmin}{arg\,min}

\title{The Best of Both Worlds: Combining Human and Machine Translations for Multilingual Semantic Parsing with Active Learning}

\author{  Zhuang Li$^{1}$, Lizhen Qu$^2$, \\
 {\bf Philip R. Cohen$^1$, Raj V. Tumuluri$^1$, Gholamreza Haffari$^{1}$} \\
    $^1$Openstream.ai, $^2$Monash University\\
   \texttt{\{zhuang.li, phil.cohen, raj, reza.haffari\}@openstream.com} \\
  \texttt{lizhen.qu@monash.edu}}

\begin{document}
\maketitle
\begin{abstract}

Multilingual semantic parsing aims to leverage the knowledge from the high-resource languages to improve low-resource semantic parsing, yet commonly suffers from the data imbalance problem. Prior works propose to utilize the translations by either humans or machines to alleviate such issues. However, human translations are expensive, while machine translations are cheap but prone to error and bias. In this work, we propose an active learning approach that exploits the strengths of both human and machine translations by iteratively adding small batches of human translations into the machine-translated training set. Besides, we propose novel aggregated acquisition criteria that help our active learning method select utterances to be manually translated. Our experiments demonstrate that an ideal utterance selection can significantly reduce the error and bias in the translated data, resulting in higher parser accuracies than the parsers merely trained on the machine-translated data. 

\end{abstract}

\section{Introduction}
\label{sec:intro}

Multilingual semantic parsing allows a single model to convert natural language utterances from multiple languages into logical forms (LFs). Due to its wide applications in various research areas, e.g. multilingual question answering and multilingual virtual assistant, multilingual semantic parsing has drawn more attention recently~\cite{zou2018learning,sherborne2020bootstrapping,li2021mtop}.

Training a multilingual semantic parser (MSP) requires training data from all target languages. However, there is a severe imbalance of data availability among languages for current multilingual semantic parsing research. The utterances in most current semantic parsing datasets are in English, while non-English data is scarce.

To overcome the data imbalance issue, prior studies translate utterances in the MSP datasets from high-resource languages (e.g. English) to the target low-resource languages of interest by either human translators~\cite{susanto2017neural,duong2017multilingual,li2021mtop} or automatic machine translation (MT) ~\cite{moradshahi2020localizing,sherborne2020bootstrapping}. Unfortunately, human translation (HT), though effective, is cost-intensive and time-consuming. While the cost of MTs is much lower than that of HTs, the low quality of the machine-translated utterances severely weakens the performance of the MSPs in the target languages.

We observe that the quality of MTs is lower than that of HTs, mainly due to translation bias and errors. First, MT systems are  likely to be influenced by algorithmic bias. Hence, the outputs of MT systems are generally less lexically and morphologically diverse than human translations~\cite{vanmassenhove2021machine}. So, there is a lexical distribution discrepancy between the machine-translated and the human-generated utterances. Second, MT systems are prone to generate translations with  errors~\cite{daems2017identifying}.

Prior study~\cite{moradshahi2020localizing} demonstrates that adding only a small portion of human-translated data into the complete set of machine-translated training data significantly improves the MSP performance on the test set of the target language. Given this observation, we propose a novel annotation strategy based on active learning (AL) that benefits from both \textbf{H}uman translations and \textbf{A}utomatic machine \textbf{T}ranslations (\hatt). It initially machine-translates all utterances in training sets from the high-resource languages to target languages. Then, for each iteration, \hatt selects a subset of utterances from the original training set to be translated by human translators, followed by adding the HT data to the MT training data. The multilingual parser is trained on the combination of both types of translated data.

We further investigate how \hatt can select utterances whose HTs maximally benefit the parser performance. We assume the performance improvement is ascribed to the less biased and erroneous training set in a mixture of the MT and HT data. We have found that resolving the bias and error issues for the translations of the most semantically diversified and representative utterances improves the parser performance to the greatest extent. Given this assumption, we provide an \textbf{A}ggregated acquisition function that scores the utterances on how much their HTs can mitigate the \textbf{B}ias and \textbf{E}rror issues for learning the multilingual parsers (\amsp). It aggregates four individual acquisition functions, two of which measure the error and bias degree for the translations of the source utterances. The other two encourage the selection of the most representative and semantically diversified utterances.

Our key contributions are as follows:

\begin{itemize}
 \item We propose a novel AL procedure, \hatt, that benefits from two popular annotation strategies for training the MSP. \hatt greatly boosts the performance of the parser trained on MT data while it requires only a small extra human annotation cost. With only 16\% of total utterances translated by humans, the parser accuracies on the multilingual \geo~\cite{susanto2017neural} and \nlmap~\cite{haas2016} test sets can be improved by up to 28\% and 5\%, respectively, compared to the accuracies of those trained on machine-translated data, and are only up to 5\% away from the \oracle parsers trained on all human data.
 
 \item We propose an aggregated acquisition function, coined \amsp, specifically designed to select utterances where their HTs mitigate translation bias and error for learning a good MSP. Compared to other SOTA acquisition baselines, given the same selection budget, our experiments consistently show \amsp consistently results in the less biased and erroneous training sets and higher parser accuracies on the multilingual \geo and \nlmap test sets.
\end{itemize}

\section{Related Work}
\label{sec:related}
\paragraph{Multilingual Semantic Parsing.} Multilingual semantic parser is an emerging field that parses utterances from multiple languages using one model.  Almost all the current MSP data are obtained by translating the utterances in existing semantic parsing datasets in the high-resource languages by the automatic translation services~\cite{moradshahi2020localizing,sherborne2020bootstrapping} or human translators~\cite{susanto2017neural,duong2017multilingual,li2021mtop,li2023active}. They don't consider conventional data collection strategies~\cite{wang2015building} for monolingual semantic parsing as they require expert knowledge in LFs, which is more expensive than bilingual knowledge. Therefore, our work follows the same strategies to leverage the knowledge from high-resource to low-resource languages. ~\citet{moradshahi2020localizing} tries to mix the human-translated data with machine-translated data to improve the parser accuracies. However, their work is only in a supervised learning setting, while our work studies how to iteratively collect utterances in an AL scenario.

\paragraph{Active Learning.} AL is to select the most valuable unlabeled instances to be annotated in order to maximize the model's performance and hence reduce the annotation cost for data-hungry machine learning models. AL has been used to MT~\citep{DBLP:conf/acl/HaffariS09}, sequence labelling~\cite{vu2019learning}, text classification~\citep{mccallum1998employing,vu2023active}, and semantic parsing~\citep{duong2018active,ni2020merging,li2023active}. Following most deep learning AL methods~\cite{duong2018active,ni2020merging,li2023active}, our work also adopts a pool-based query strategy, which means we sample batches from a large pool of unlabelled data instead of evaluating examples one by one from an incoming stream. Among all the AL for semantic parsing works,~\citet{li2023active} is the one most similar to ours, which selects utterances to be translated. However, they do not utilize MT systems.

\section{Multilingual Semantic Parsing with Automatic Machine Translation}
\label{sec:msp}

An MSP is a parametric model $P_{\theta}(\vy|\vx)$ that maps a natural language utterance $\vx \in \mathcal{X}$ into a formal meaning representation $\vy \in \mathcal{Y}$, where $\mathcal{X} = \bigcup_{l \in L} \mathcal{X}_l$ includes utterances in different languages $L$. The standard training objective for a multilingual parser is,
\begin{equation}
    \argmax_{\theta} \prod_{\vx, \vy \in \mathcal{D}_{L}} P_{\theta}(\vy|\vx)
\end{equation}
\noindent where $\mathcal{D}_{L} = \bigcup_{l \in L} \mathcal{D}_{l}$ includes training data where utterances are from multiple languages $L$. 
\subsection{Difficulties for Multilingual Semantic Parsing Utilizing Machine Translation}
Although using an MT system to train an MSP is cost-effective, the parser performance is usually much lower than the one trained with human-translated data. For example, as shown in Table~\ref{tab:bias_analysis}, the parsers trained on HTs all have significantly higher accuracies than those trained on MTs in different  settings. Such performance gaps are due to two major issues of the MT data, discussed below.

\paragraph{Translation Bias.}
\begin{table}[t]
    \vspace{-2mm}
\centering
  \resizebox{\textwidth}{!}{%
  \begin{tabular}{|c||cc| cc| cc|cc|}
    \toprule
    \multirow{2}{*}{Metrics} &
      \multicolumn{2}{c|}{\geode} &
      \multicolumn{2}{c|}{\geoth} &
      \multicolumn{2}{c|}{\geoel} &
      \multicolumn{2}{c|}{\nlmapde}\\
     & HT& MT & HT& MT & HT& MT & HT & MT  \\
      \hline \hline
       Accuracy$\uparrow$ &  78.14 & 47.21 & 79.29 & 56.93 & 80.57 & 68.5 & 81.57 &  67.86 \\
             \hline
           \hline
 BT Discrepancy Rate$\downarrow$ &  2\% & 11\% & 3\% & 12\% & 3\% & 10\% & 2\% &  10\% \\
      \hline
    \hline
          JS$\downarrow$ &  36.67 & 59.95 & 32.02 & 73.83 & 33.67 & 56.36 & 33.78 &  46.84 \\
    MAUVE$\uparrow$ & 96.01 & 22.37  & 97.52  &8.48 & 97.12 & 45.01 & 97.34 & 70.24 \\
\hline
MTLD$\uparrow$ & 26.02&  22.50  & 20.74  & 19.07 & 28.16& 27.08 & 44.80&42.38       \\
    \bottomrule
  \end{tabular}%
  }
    \caption{The scores of five metrics to measure the quality of the HTs and MTs in German (De), Thai (Th) and Greek (El) of the utterances in \geo and \nlmap. $\uparrow$/$\downarrow$ means the higher/lower score the better. See \textbf{Evaluation} in Sec.~\ref{sec:experiments} for the details of Accuracy, MTLD, JS, MAUVE and \btmetric. 
    \vspace{-3mm}}
  \label{tab:bias_analysis}
  \vspace{-2mm}
\end{table}
Many existing MT systems amplify biases observed in the training data~\cite{vanmassenhove2021machine}, leading to two problems that degrade the parsers' performance trained on MT data:
\begin{itemize}
    \item The MTs lack lexical diversity~\cite{vanmassenhove2021machine}. 
As shown in Table \ref{tab:bias_analysis}, MTLD~\cite{vanmassenhove2021machine} values show that the HTs of utterances in multilingual \geo and \nlmap are all more lexically diversified than MTs. Several studies~\cite{shiri2022paraphrasing,xu2020autoqa,wang2015building,zhuo2023robustness,huang2021robustness} indicate that lexical diversity of training data is essential to improving the generalization ability of the parsers.

\item The lexical distribution of the biased MTs is different to the human-written text. 
The two metrics, Jensen–Shannon (JS) divergence~\cite{manning1999foundations} and MAUVE~\cite{pillutla2021mauve}, in Table~\ref{tab:bias_analysis} show the HTs of utterances in \geo and \nlmap are more lexically close to the human-generated test sets than MTs.
\end{itemize}

\paragraph{Translation Error.} 
MT systems often generate translation errors due to multiple reasons, such as underperforming MT models or an absence of contextual understanding~\cite{wu2023document,wu4330827improving}, leading to discrepancies between the source text and its translated counterpart. One common error type is mistranslation~\cite{vardaro2019translation}, which alters the semantics of the source sentences after translation. Training an MSP on the mistranslated data would cause incorrect parsing output, as LFs are the semantic abstraction of the utterances. 
BT Discrepancy Rate in Table~\ref{tab:bias_analysis} demonstrates the mistranslation problem is more significant in the machine-translated datasets.

\section{Combining Human and Automatic Translations with Active Learning}
\label{sec:method}

To mitigate the negative effect of translation bias and error in the MT data, we propose \hatt, which introduces extra human supervision to machine supervision when training the MSPs. Two major intuitions motivate our training approach: 
\begin{itemize}
\item Adding the HTs to the training data could enrich its lexical and morphological diversity and ensure that the lexical distribution of the training data is closer to the human test set, thus improving the parsers' generalization ability~\cite{shiri2022paraphrasing,xu2020autoqa,wang2015building}. 
\item HTs are less erroneous than MTs~\cite{freitag2021experts}. The parser could learn to predict correct abstractions with less erroneous training data.
\end{itemize}

Our \hatt AL setting considers only the \textit{bilingual} scenario. One of the languages is in high-resource, and the other one is in low-resource. However, it is easy to extend our method to more than two languages. We assume access to a well-trained black-box multilingual MT system, $g^{mt}(\cdot)$, and a semantic parsing training set that includes utterances in a high-resource language $l_s$ (e.g. English) paired with LFs, $\mathcal{D}_s = \{(\vx_{s}^i, \vy^i)\}_{i=1}^{N}$, two human-generated test sets $\mathcal{T}_{s} = \{(\vx_{s}^i, \vy^i)\}_{i=1}^{M}$ and $\mathcal{T}_{t} = \{(\vx_{t}^i, \vy^i)\}_{i=1}^{M}$ with utterances in high and low-resource languages, respectively. Each utterance $\vx_{s}$ in $\mathcal{D}_s$ is translated into the utterance $\hat{\vx}_{t} = g^{mt}_{s\rightarrow t}(\vx_{s})$ in the target language $l_t$ by the MT system, $\hat{\mathcal{D}}_{t} = \{(\hat{\vx}^i_{t}, \vy^i)\}_{i=1}^{N}$. The goal of our AL method is to select an optimal set of utterances from the training data in the source language, $\tilde{\mathcal{D}}_s \in \mathcal{D}_s$, and ask human translators to translate them into the target language, denoted by $\bar{\mathcal{D}}_t = g^{ht}_{s\rightarrow t}(\tilde{\mathcal{D}}_s)$, for training a semantic parser on the union of $\bar{\mathcal{D}}_t$ and $\hat{\mathcal{D}}_{t}$. 
The selection criterion is based on the \textit{acquisition functions} that score the source utterances. Following the conventional batch AL setting~\cite{duong2018active}, there are $Q$ selection rounds. At the $q$th round, AL selects utterances with a budget size of $K_q$.

The detailed \hatt AL procedure iteratively performs the following steps as in Algorithm.~\ref{algo:al}.

\begin{algorithm}[ht]
{\small
\SetKwData{Left}{left}\SetKwData{This}{this}\SetKwData{Up}{up}
\SetKwFunction{Union}{Union}\SetKwFunction{FindCompress}{FindCompress}
\SetKwInOut{Input}{Input}\SetKwInOut{Output}{Output}
\SetAlgoLined
\Input{Initial training set $\mathcal{D}^{0}=\mathcal{D}_s \cup \hat{\mathcal{D}}_t$, source utterance pool $\mathcal{D}_s$, budget size $K_q$, number of selection rounds $Q$, human annotators $g^{ht}(\cdot)$}
\Output{A well-trained multilingual parser $P_{\theta}(\vy|\vx)$}

\textcolor{blue}{\# \textit{Train the initial parser on the initial data}} \\
     Update $\theta$ of $P_{\theta}(\vy|\vx)$ with $\nabla_{\theta}\mathcal{L}(\theta)$ on $\mathcal{D}^{0}$ \\
Evaluate $P_{\theta}(\vy|\vx)$ on $\mathcal{T}_s$ and $\mathcal{T}_t$ \\
Estimate the acquisition function $\phi(\cdot)$\\
$\bar{\mathcal{D}}^{0}_t = \emptyset$ \textcolor{blue}{\# \textit{Empty set of human-translated data}} \\
$\bar{\mathcal{D}}^{0}_s = \mathcal{D}_s$ \textcolor{blue}{\# \textit{Initial source utterance pool}}\\
\For{$q \gets 1$ to $Q$} 
{
\textcolor{blue}{\# \textit{Select a subset $\tilde{\mathcal{D}}^{q}_s \in \mathcal{D}^{q-1}_s$ of the size $K_q$ with the highest scores ranked by the acquisition function $\phi(\cdot)$}} \\

$\tilde{\mathcal{D}}^{q}_s = \text{TopK} (\phi(\bar{\mathcal{D}}^{q-1}_s), K_q)$ \\
$\bar{\mathcal{D}}^{q}_s = \bar{\mathcal{D}}^{q-1}_s \setminus \tilde{\mathcal{D}}^{q}_s$ \\
\textcolor{blue}{\# \textit{Translate the utterances in $\tilde{\mathcal{D}}^{q}_s$ into the target language $l_t$ by human annotators}} \\
$\mathcal{D}^{q}_t = g^{ht}_{s\rightarrow t}(\tilde{\mathcal{D}}^{q}_s)$\\
\textcolor{blue}{\# \textit{Merge all human-translated data}} \\
$\bar{\mathcal{D}}^{q}_t = \bar{D}^{q-1}_t \cup D^{q}_{t}$  \\
\textcolor{blue}{\# \textit{Add the human-translated data into the training data}} \\
$\mathcal{D}^{q} = \mathcal{D}_s \cup \hat{\mathcal{D}}_t \cup \bar{\mathcal{D}}^{q}_t$ \\
\textcolor{blue}{\# \textit{Train the parser on the updated data}} \\
Update $\theta$ of $P_{\theta}(\vy|\vx)$ with $\nabla_{\theta}\mathcal{L}(\theta)$ on $\mathcal{D}^{q}$ \\
Evaluate $P_{\theta}(\vy|\vx)$ on $\mathcal{T}_s$ and $\mathcal{T}_t$\\
Re-estimate  $\phi(\cdot)$
}
}
\caption{\hatt procedure
}
\label{algo:al}
\end{algorithm}

\subsection{Acquisition Functions} 
\label{sec:acq}

The acquisition functions assign higher scores to those utterances whose HTs  can boost the parser's performance more than the HTs of the other utterances.
The prior AL works~\cite{sener2018active,zhdanov2019diverse,nguyen2004active} suggest that the most representative and diversified examples in the training set improve the generalization ability of the machine learning models the most. Therefore, we provide a hypothesis that \textit{we should select the representative and diversified utterances in the training set, whose  current translations have significant bias and errors}. We postulate fixing problems of such utterances improves the parsers' performance the most. We derive four acquisition functions based on this hypothesis to score the utterances. Then, \amsp aggregates these acquisition functions to gain their joint benefits. In each AL round, the utterances with the highest \amsp scores are selected.

\paragraph{Translation Bias.} We assume an empirical conditional distribution, $P_{e}^{q}(\vx_t|\vx_s)$, for each utterance $\vx_s$ in $\mathcal{D}_{s}$ at $q$th AL selection round. 
Intuitively, the $\vx_s$ with the most biased translations should be the one with the most skewed empirical conditional distribution. Therefore, we measure the translation bias by calculating the entropy of the empirical conditional distribution, $H(P_{e}^{q}(\vx_{t}|\vx_s))$, and select the $\vx_s$ with the lowest entropy. Since the translation space  $\mathcal{X}_{t}$ is exponentially large, it is intractable to directly calculate the entropy. Following ~\cite{settles-craven-2008-analysis}, we adopt two approximation strategies, \textit{N-best Sequence Entropy} and \textit{Maximum Confidence Score}, to approximate the entropy.

\noindent\textit{$\bullet$ N-best Sequence Entropy:}

\begin{align}
    \phi_b(\vx_s) = - \sum_{\hat{\vx}_t \in \mathcal{N}} \hat{P_{e}^{q}}(\hat{\vx}_t|\vx_s) \log \hat{P_{e}^{q}}(\hat{\vx}_t|\vx_s)
\end{align}

\noindent\ignorespaces where $\mathcal{N} = \{\hat{\vx}^{1}_t,...,\hat{\vx}^{N}_t\}$ are the $N$-best hypothesis sampled from the empirical distribution $P_{e}^{q}(\vx_{t}|\vx_s)$. $\hat{P_{e}^{q}}(\hat{\vx}_t|\vx_s)$ is re-normalized from $P_{e}^{q}(\hat{\vx}_t|\vx_s)$ over $\mathcal{N}$, which is only a subset of $\mathcal{X}_{t}$.

\noindent\textit{$\bullet$ Maximum Confidence Score (MCS)}:

\begin{align}
    \phi_b(\vx_s) & =  \log P_{e}^{q}(\vx'_t|\vx_s) \\
    s.t. \vx'_t & = \argmax_{\vx_t} P_{e}^{q}(\vx_{t}|\vx_s)
\end{align}
\noindent\ignorespaces It is difficult to obtain the empirical distribution as we know neither of the two distributions that compose the empirical distribution. Therefore, we use distillation training~\cite{hinton2015distilling} to train a translation model that estimates $P_{e}^{q}(\vx_t|\vx_s)$ on all the bilingual pairs $(\vx_s, \vx_t)$ in the MSP training data $\mathcal{D}^{q}$. Another challenge is that $\mathcal{D}^{q}$ is too small to distil a good translation model that imitates the mixture distribution. Here, we apply a bayesian factorization trick that factorizes $P_{e}^{q}(\vx_{t}|\vx_s) = \sum_{\vy \in \mathcal{Y}} P_{e}^{q}(\vx_t|\vy) P_{e}^{q}(\vy|\vx_{s})$, where $\vy$ ranges over LFs representing the semanics. As there is a deterministic mapping between $\vx_s$ and the LF, $P_{e}^{q}(\vy|\vx_{s})$ is an one-hot distribution. Thus, we only need to estimate the entropy, $H(P^{q}_{e}(\vx_t|\vy))$. This has  a nice intuition:  the less diversified data has less lexically diversified utterances per each LF. Note that if we use this factorization, all $\vx_s$ that share the same LF have the same scores. 


We use the lightweight, single-layer, recurrent neural network-based Seq2Seq model to estimate $P_{e}^{q}(X_t|\vx_s)$ or $P^{q}_{e}(\vx_t|\vy)$. It only takes approximately 30 seconds to train the model on \geo.
Ideally, every time a new source utterance $\vx_s$ is selected, $P_{e}^{q}(\vx_t|\vx_s)$ should be re-estimated. However, we only re-estimate $P_{e}^{q}(\vx_t|\vx_s)$ once at the beginning of each selection round to reduce the training cost.

\paragraph{Translation Error.} 
Similar to~\citet{haffari-etal-2009-active}, we leverage back-translations (BTs) to measure the translation error. We conjecture that if the translation quality for one source utterance $\vx_s$ is good enough, the semantic parser should be confident in the LF of the source utterance conditioned on its BTs. 
Therefore, we measure the translation error for each $\vx_s$ as the expected parser's negative log-likelihood in its corresponding LF $\vy_{\vx_s}$ over all the BTs of $\vx_s$, $\mathbb{E}_{P^{q}_{e}(\vx_{t}|\vx_s)}[-\log(P^{q}_{\theta}(\vy_{\vx_s}|g_{t \rightarrow s}^{mt}(\vx_t)))]$, where $P^{q}_{\theta}$ is the parser trained at $q$th round. To approximate the expectation, we apply two similar strategies as mentioned in \textit{Translation Bias}.

\noindent\textit{$\bullet$ N-best Sequence Expected Error:}

\begin{small}
\begin{align}
    \phi_e(\vx_s) & = -\sum_{\hat{\vx}_t \in \mathcal{N}_{\vy_{\vx_s}}} \hat{P}_{e}^{q}(\hat{\vx}_t|\vx_s) \log P_{\theta}(\vy_{\vx_s}|g_{t \rightarrow s}^{mt}(\vx_t)) 
\end{align}
\end{small}

\noindent\ignorespaces where $\mathcal{N}_{\vy_{\vx_s}}$ is the set of translations in $\mathcal{D}^q$ that share the same LF $\vy_{\vx_s}$ with $\vx_s$. We only back-translate utterances in $\mathcal{D}^q$ to reduce the cost of BTs.

\noindent\textit{$\bullet$  Maximum Error:}

\begin{small}
\begin{align}
    \phi_e(\vx_s) & = - \log P^{q}_{\theta}(\vy_{\vx_s}|g^{mt}_{t \rightarrow s}(\vx'_t)) \\
        s.t. \vx'_t & = \argmax_{\vx_t} P^{q}_{e}(\vx_{t}|\vx_s)
\end{align}
\end{small}

\noindent\ignorespaces We use the same distilled translation model $P^{q}_{e}(\vx_{t}|\vx_s)$ used in \textit{Translation Bias}.

\paragraph{Semantic Density.}
The previous AL works~\cite{nguyen2004active,donmez2007dual} have found that the most \textit{representative} examples improve the model performance the most. Therefore we desire to reduce the translation error and bias for the translations of the most representative source utterances. As such,  the utterances should be selected from the dense regions in the semantic space,
\begin{equation}
    \phi_s(\vx_s) = \log P(\vx_s).
\end{equation}
We use kernel density estimation~\cite{botev2010kernel} with the exponential kernel to estimate $P(\vx_s)$, while other density estimation methods could be also used. The feature representation of $\vx_s$ for density estimation is the average pooling of the contextual sequence representations from the MSP encoder. The density model is re-estimated at the beginning of each query selection round.

\paragraph{Semantic Diversity.} The role of the semantic diversity function is twofold. First, it prevents the AL method from selecting similar utterances. Resolving the bias and errors of similar utterances in a small semantic region does not resolve the training issues for the overall dataset. Second, semantic diversity correlates with the lexical diversity, hence improving it also enriches lexical diversity.
\begin{align}
    \phi_d(\vx_s) =     
    \begin{cases}
      0 & \text{if $c(\vx_s) \notin \bigcup_{\vx^{i}_s \in \mathcal{S}} c(\vx^{i}_s)$}\\
      -\infty & \text{Otherwise}
    \end{cases}
\end{align}
\noindent\ignorespaces where $c(\vx_s)$ maps each utterance $\vx_s$ into a cluster id and $\mathcal{S}$ is the set of cluster ids of the selected utterances. We use a clustering algorithm to diversify the selected utterances as in~\cite{ni2020merging,nguyen2004active}. The source utterances are partitioned into $|\mathcal{C}|$ clusters. We select one utterance at most from each cluster. Notice the number of clusters should be greater than or equal to the total budget size until current selection round, $|\mathcal{C}| \geq \sum^{q}_{i=1}K_{i}$. The clusters are re-estimated every round. To ensure the optimal exploration of semantic spaces across different query rounds, we adopt Incremental K-means~\cite{liu2020active} as the clustering algorithm. At each new round, Incremental K-means considers the selected utterances as the fixed cluster centres, and learns the new clusters conditioned on the fixed centres. The feature representation of $\vx_s$ for Incremental K-means is from MSP encoder as well.

\paragraph{Aggregated Acquisition.}
We aggregate the four acquisition functions into one,
\begin{align}\nonumber
\label{eq:sample}
    & \phi_{A}(\vx_s) = \sum_{k}\alpha_{k} \phi_{k}(\vx_s) 
\end{align}
\noindent where $\alpha_{k}$'s are the coefficients. Each $\phi_{k}(\vx_s)$ is normalized using quantile normalization~\cite{bolstad2003comparison}. Considering the approximation strategies we employ for both \textit{Translation Bias} and \textit{Translation Error}, \amsp can be denoted as either \amspnbest or \amspmax. The term \amspnbest is used when we apply \textit{N-best Sequence Entropy} and \textit{N-best Sequence Expected Error}. On the other hand, \amspmax is used when we implement \textit{Maximum Confidence Score} and \textit{Maximum Error} strategies.

\section{Experiments}
\label{sec:experiments}
\paragraph{Datasets.}
We evaluate our AL method for MSP on two datasets, \geo~\cite{susanto2017neural} and \nlmap~\cite{haas2016} with multilingual human-translated versions. \geo includes 600 utterances-LF pairs as the training set and 280 pairs as the test set. \nlmap includes 1500 training examples and 880 test examples. 

In our work, we consider English as the \textit{resource-rich} source language and use Google Translate System\footnote{https://translate.google.com/} to translate all English utterances in \geo into German (De), Thai (Th), Greek (El) and the ones in \nlmap into German, respectively. The AL methods actively sample English utterances, the HTs of which are obtained from the multilingual \geo and \nlmap.

\paragraph{Active Learning Setting.} The \hatt active learning procedure performs five rounds of query, which accumulatively samples 1\%,  2\%, 4\%, 8\% and 16\% of total English utterances in \geo and \nlmap. We only perform five rounds as we found the performance of the multilingual parser is saturated after sampling 16\% of examples with most acquisition functions.

\paragraph{Base Parser.} We use \bertlstm as our multilingual parser~\cite{moradshahi2020localizing}. It is a Seq2Seq model with the copy mechanism~\cite{gu2016incorporating} that applies Multilingual BERT-base~\cite{DBLP:journals/corr/abs-1810-04805} as the encoder and LSTM~\cite{hochreiter1997long} as the decoder.

\paragraph{Baselines.} We compare \amsp with eight acquisition baselines and an oracle baseline. 
\begin{enumerate}
\item \textbf{Random} randomly selects English utterances in each round. 
\item \textbf{Cluster}~\cite{ni2020merging,li2021total} partitions the utterances into different groups using K-means and randomly selects one example from each group. 

\item \textbf{LCS (FW)}~\cite{duong2018active} selects English utterances for which the parser is least confident in their corresponding LFs, $\vx = \argmin_{\vx}p_{\theta}(\vy|\vx)$. 

\item \textbf{LCS (BW)}~\cite{duong2018active}, on the opposite of LCS (BW), trains a text generation model to generate text given the LF. The English utterances are selected for which the text generation model is least confident conditioned on their corresponding LFs, $\vx = \argmin_{\vx}p_{\theta}(\vx|\vy)$. 

\item \textbf{Traffic}~\cite{sen2020uncertainty} selects utterances with the lowest perplexity and highest frequency in terms of their corresponding LFs. 

\item \textbf{CSSE}~\cite{hu2021phrase} combines the density estimation and the diversity estimation metrics to select the most representative and semantically diversified utterances. 

\item \textbf{RTTL}~\cite{haffari-etal-2009-active,DBLP:conf/acl/HaffariS09} uses BLEU~\cite{papineni2002bleu} to estimate the translation information losses between the BTs and the original utterances to select utterances with  highest losses. 

\item \textbf{LFS-LC-D}~\cite{li2023active} is the selection method for MSP, which enriches the diversity of lexicons and LF structures in the selected examples. 

\item \textbf{ORACLE} trains the parser on the combination of English data, machine-translated data, and the complete set of human-translated data.
\end{enumerate}

\paragraph{Evaluation.} We evaluate the AL methods by measuring the accuracy of the MSP, the bias of the training set, and the semantic discrepancy rate between the selected utterances and their BTs.
\begin{itemize}

\item \textbf{Accuracy:} To evaluate the performance of the MSP, we report the accuracy of exactly matched LFs as in ~\cite{dong2018coarse} at each query round. As the parser accuracies on the English test sets are not relevant to evaluating the active learning method, we only report the accuracies on the test sets in the \textit{target} languages. See Appendix~\ref{app:english_res} for the English results.

\item \textbf{Bias of the Training Set:} We use three metrics to measure the bias of the training data in the target language at each query round. 

\begin{enumerate}

\item \textit{Jensen–Shannon (JS) divergence}~\cite{pillutla2021mauve} measures the JS divergence between the n-gram frequency distributions of the utterances in the training set $\hat{\mathcal{D}}_t \cup \bar{\mathcal{D}}^{q}_t$ generated by each AL method and test set $\mathcal{T}_{t}$. 

\item \textit{MAUVE}~\cite{pillutla2021mauve} compares the learnt distribution from the training set to the distribution of human-written text in the test set $\mathcal{T}_{t}$ using Kullback–Leibler divergence~\cite{kullback1951information} frontiers. Here we use n-gram lexical features from the text when calculating MAUVE. \textit{JS} and \textit{MAUVE} together measure how lexically "human-like" the generated training set is. 
 
\item \textit{MTLD}~\cite{mccarthy2005assessment} reports the mean length of word strings in the utterances in $\hat{\mathcal{D}}_t \cup \bar{\mathcal{D}}^{q}_t$ that maintain a given TTR~\cite{templin1957certain} value, where TTR is the ratio of different tokens to the total number of tokens in the training data. \textit{MTLD} evaluate the lexical diversity of the training set. 

\end{enumerate}

\item \textbf{BT Discrepancy Rate:} Since we do not possess bilingual knowledge, we use BT to access the translation quality~\cite{tyupa2011theoretical}. At each query round, we randomly sample 100 utterances from the utterances selected by each acquisition in 5 seeds' experiments. The BT is obtained by using Google Translation to translate the MTs of the 100 sampled utterances back to English. Two annotators manually check the percentage of the BTs which are not semantically equivalent to their original utterances. We only consider a BT discrepancy when both annotators agree. Ideally, the utterances with fewer mistranslations would see fewer semantic discrepancies between the BTs and the original.
\end{itemize}

\subsection{Main Results and Discussion.}
\begin{figure*}[ht!]
           \vspace{-2mm}
    \centering
    \includegraphics[width=\textwidth]{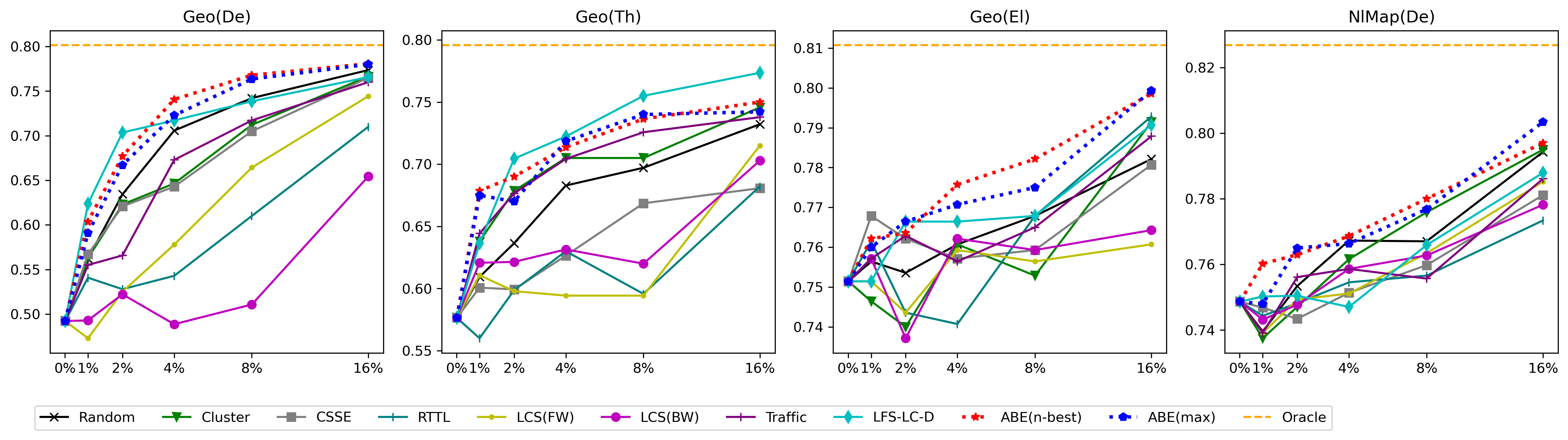}
    \caption{The parser accuracies at different query rounds using various acquisitions on the test sets of \geode, \geoth, \geoel and \nlmapde. Orange dash lines indicate the accuracies of \oracle multilingual parsers. All experiments are run 5 times with a different seed for each run.}
    \label{fig:main_al_result}
\end{figure*}

\paragraph{Effectiveness of \hatt.} Fig.~\ref{fig:main_al_result} shows that \hatt significantly improves the parser accuracies on all test sets by adding only a small amount of HTs into the machine-translated training set. For example, with 16\% of English utterances translated by humans, \hatt improves the parser accuracies by up to 28\% and 25\%, respectively, on \geode and \geoth test sets. On the other hand, on \geoel and \nlmapde test sets, the accuracy improvement by \hatt is only up to 5\% because the parser has already achieved a decent performance after it is trained on the MT data. According to Table~\ref{tab:bias_analysis}, we speculate that the training sets of \geoel and \nlmapde are less biased than those of \geoth and \geode. Overall for all dataset settings, if we apply \hatt with \amsp, the multilingual parsers can perform comparably to the \oracle parsers with no more than 5\% differences in terms of accuracies at an extra expense of manual translating 16\% of English utterances.

\paragraph{Effectiveness of \amsp.} The \amsp method has been demonstrated to consistently achieve superior performance over the baselines by utilizing a combination of four important measurements. In contrast, the acquisition baselines focus on only one or two of these measurements, and as a result, they fail to effectively address issues of bias and error across all datasets and languages. Despite this, these baselines may still perform well in certain specific settings, such as \lfslcd performing slightly better than \amsp on the \geoth dataset. However, it should be noted that this performance is not consistent across all settings. Three baselines, \lcsfw, \lcsbw, and \rttl, consistently perform lower than the others. \lcsfw tends to select similar examples, which lack semantic diversity. \rttl is designed to choose the utterances with the most erroneous translations, while such utterances are mostly the tail examples given our inspection. \amsp overcomes this issue by balancing the \textit{Translation Error} term with the \textit{Semantic Density}. \lcsbw has an opposite objective with our \textit{Translation Bias}, which means it selects the utterances with the most translation bias. Therefore, though \lcsbw performs well in the AL scenario in~\citet{duong2018active} for semantic parsing, it performs worst in our scenario.

\paragraph{Bias, Error and Parser Performance.} As in Table~\ref{tab:main_diversity_error}, we also measure the bias of the training set and the BT discrepancy rates of the selected utterances at the final selection round for each acquisition function. We can see that the parser accuracy directly correlates with the training set's bias degree. The bias of the training set acquired by \rand, \traffic and \cluster, \lfslcd, and \amsp score better in general than the other baselines in terms of the bias metrics, resulting in a better parser accuracy. \rttl and \lcsfw that select utterances with more erroneous translations do not necessarily guarantee better accuracies for parsers. Our following ablation study shows that the parser performance can be improved by correcting the translation errors for the most representative utterances.

\begin{table*}[ht]
\centering
  \resizebox{\textwidth}{!}{%
  \begin{tabular}{|c|c|cccccccc|cc|c|}
    \toprule
     Metric & No\ HT & \rand & \cluster & \csse & \rttl & \lcsfw & \lcsbw & \traffic & \lfslcd & \amspnbest & \amspmax & \oracle  \\
      \midrule
          BT Discrepancy Rate & - & 11\%  & 14\% & 11\% & 21\% & \textbf{22\%} & 8\% & 14\% & 10\% & 17\% & 18\% & - \\
     \hline
          \hline
    JS$\downarrow$ & 59.95 &  54.15 & 54.71 & 55.53 & 54.56 & 54.38 & 56.13 & 54.58 & 54.26 &54.16 & \textbf{53.97} & 45.12 \\
    MAUVE$\uparrow$ & 22.37 &  36.99 & 36.12 & 34.52 & 35.53 & 31.61 & 29.75 & 35.67 & 36.87 &\textbf{38.96} & 35.13 & 73.04 \\
\hline
\hline
  MTLD$\uparrow$ & 22.50 & 23.79 &  23.32  & 22.65  & 22.89 & 23.00 & 22.27& 23.42 & \textbf{23.97}&23.80 & 23.78  & 24.23 \\
          \hline
    \bottomrule
  \end{tabular}
  }
    \caption{Using different acquisitions at the final query round, we depict the scores of the metrics to measure the bias of the training sets in \geode and the BT discrepancy rates of the total selected utterances. 
      }
  \label{tab:main_diversity_error}
\end{table*}

\begin{figure}[ht!]
    \centering
    \includegraphics[width=0.9\textwidth]{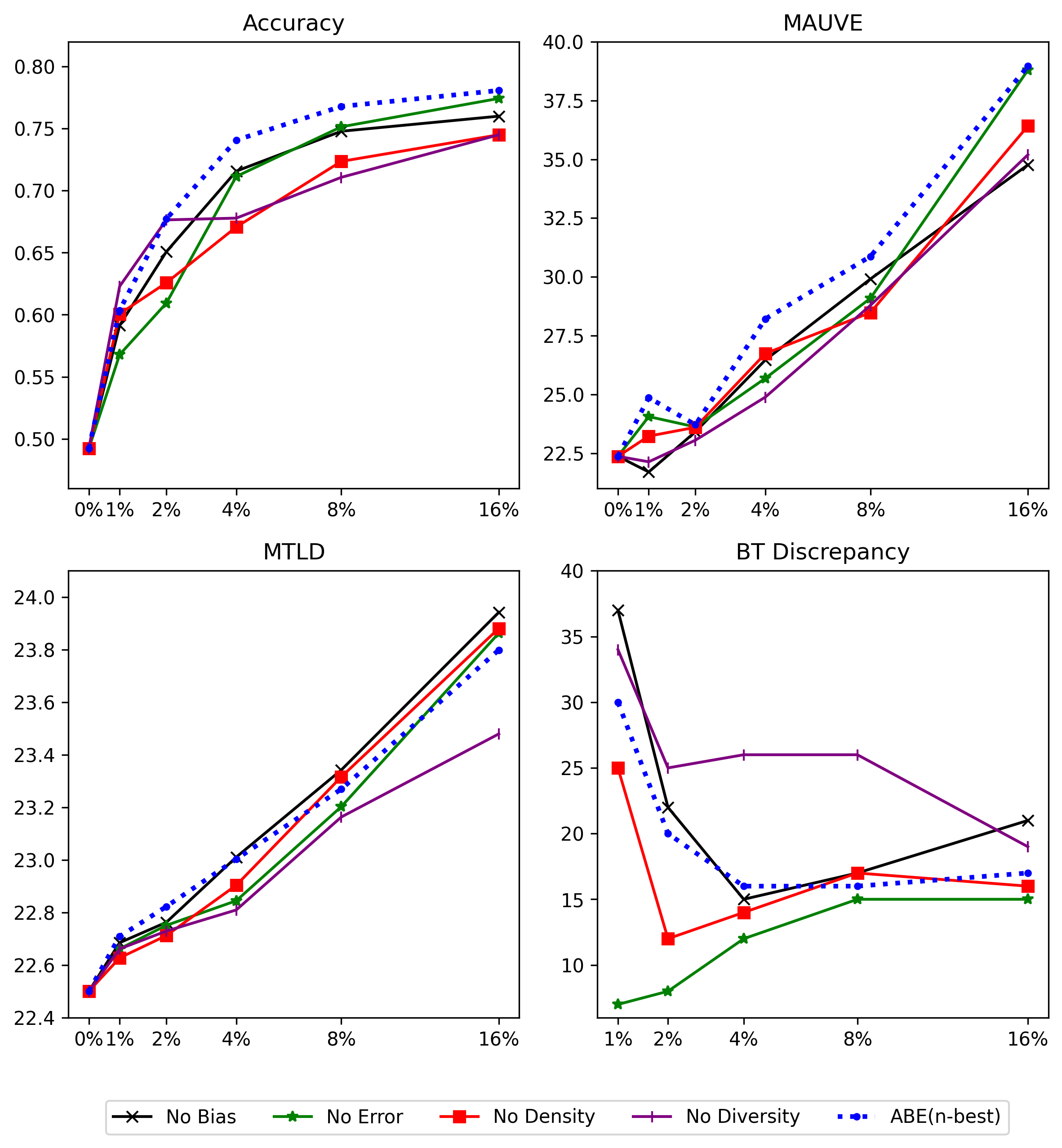}
    \caption{Using \amspnbest with each term removed, we depict the parser accuracies on the \geode test set, the MAUVE and MTLD scores of the \geode training sets and the BT Discrepancy rate of the selected utterances from English \geo at each query round.\vspace{-4mm}}
    \label{fig:remove_al_result}
\end{figure}

\subsection{Ablation Study.}
\paragraph{Influence of different Acquisition Functions.} As in Fig.~\ref{fig:remove_al_result}, we evaluate the effectiveness of each acquisition by observing how removing each acquisition from \amspnbest influences the parser performance, the bias of the training set and the BT Discrepancy rate of the selected utterances. We can see that removing all terms degrades the parser performance. However, each acquisition contributes to the parser accuracy due to different reasons.

Translation Bias and Semantic Diversity contribute to the parser performance mainly due to alleviating the bias of the training set. Excluding Translation Bias does not influence the lexical diversity, while the lexical similarity between the training and test sets becomes lower. Removing Semantic Diversity drops the lexical similarity as well. But it more seriously drops the lexical diversity when the sampling rates are high.

Removing Translation Error significantly decreases the parser accuracy and BT Discrepancy rate in the low sampling regions. However, when the selection rate increases, gaps in parser accuracies and BT Discrepancy rates close immediately. Translation Error also reduces the bias by introducing correct lexicons into the translations.

Removing Semantic Density also drops the parser performance as well. We inspect that Semantic Density contributes to parser accuracy mainly by combing with the Translation Error term. As in Appendix~\ref{app:only_term}, using Translation Error or Semantic Density independently results in inferior parser performance. We probe that Translation Error tends to select tail utterances from the sparse semantic region given the TSNE~\cite{van2008visualizing} plots at Appendix~\ref{app:rep_picture}.

\begin{figure}[ht]
    \centering
    \includegraphics[width=\textwidth]{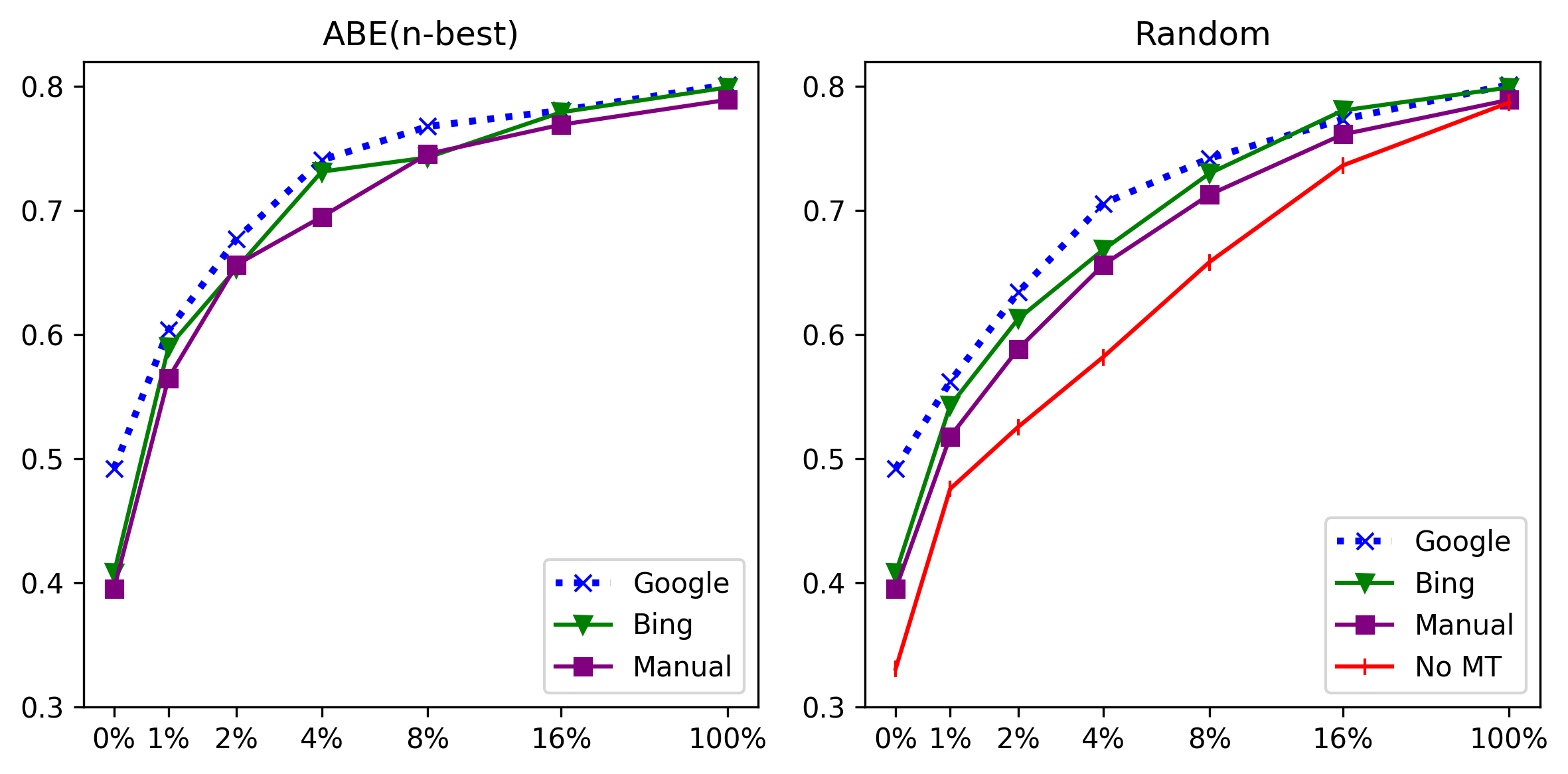}
    \caption{
The parser accuracies across various query rounds on the \geode test set. We use two selection methods: \amspnbest (shown on the left) and \rand (shown on the right). For each method, we use data from different MT systems - Google, Bing, and our bespoke manually trained MT system. This manual MT system was developed without any pre-training weight, utilizing a limited set of bilingual data. The model architecture was based on the framework proposed by~\citet{ott2018scaling}. In addition to these, we also conducted tests without utilizing any MT data.}
    \label{fig:mt_al_result}
    
\end{figure}

\paragraph{Influence of MT Systems.}
As in Fig.~\ref{fig:mt_al_result} (Right), at all query rounds, the multilingual parsers perform better with MT data in the training set, showing that MT data is essential for improving the parser's performance when a large number of HTs is not feasible. The quality of the MT data also significantly influences the parser performance when having no HT data in the training set. The accuracy difference between the parsers using Google and Bing translated data is greater than 10\% when active learning has not been performed. However, after obtaining the HT data by \hatt, the performance gaps close immediately, although the MT data of better quality brings slightly higher performance. When having all utterances translated by humans, the performance differences between parsers with different MT systems can be negligible.

Fig.~\ref{fig:mt_al_result} also demonstrates that \amspnbest outperforms \rand, a strong acquisition baseline, with all three different MT systems. \amspnbest is also more robust to the MT systems than \rand. The performance gaps for the parsers with \amspnbest are much smaller than those with \rand when applying different MT systems.

\section{Conclusion}
\label{sec:conclusion}

We have tackled the problem of data imbalance when adapting an MSP to a low-resource language. We presented methods to efficiently collect a small amount of human-translated data to reduce bias and error in the training data, assuming a realistic scenario with an MT system and budget constraints for human annotation. Our experiments show that by manually translating only 16\% of the dataset, the parser trained on this mixed data outperforms parsers trained solely on machine-translated data and performs similarly to the parser trained on a complete human-translated set.
\section*{Limitations}

One of the limitations is the selection of hyperparameters. At present, we determine the optimal hyperparameters based on the performance of the selection methods on an existing bilingual dataset. For example, to identify the appropriate utterances to be translated from English to German, we would adjust the hyperparameters based on the performance of the methods on existing datasets in English and Thai. However, this approach may not always be feasible as such a dataset is not always available, and different languages possess distinct characteristics. As a result, the process of tuning hyperparameters on English-Thai datasets may not guarantee optimal performance on English-German datasets. As a future direction, we intend to investigate and develop more effective methods for hyperparameter tuning to address this limitation.
\section*{Acknowledgement}
I would like to extend my sincere gratitude to Minghao Wu for his invaluable assistance in building the manual MT systems. I am equally grateful to both Minghao Wu and Thuy-Trang Vu for their insightful comments and suggestions during the preparation of this paper.

\bibliography{anthology,custom}
\bibliographystyle{acl_natbib}

\clearpage
\appendix
\section{Appendix}
\label{sec:appendix}
\subsection{Sensitivity Analysis}
\begin{figure}[ht]
    \centering
    \includegraphics[width=\textwidth]{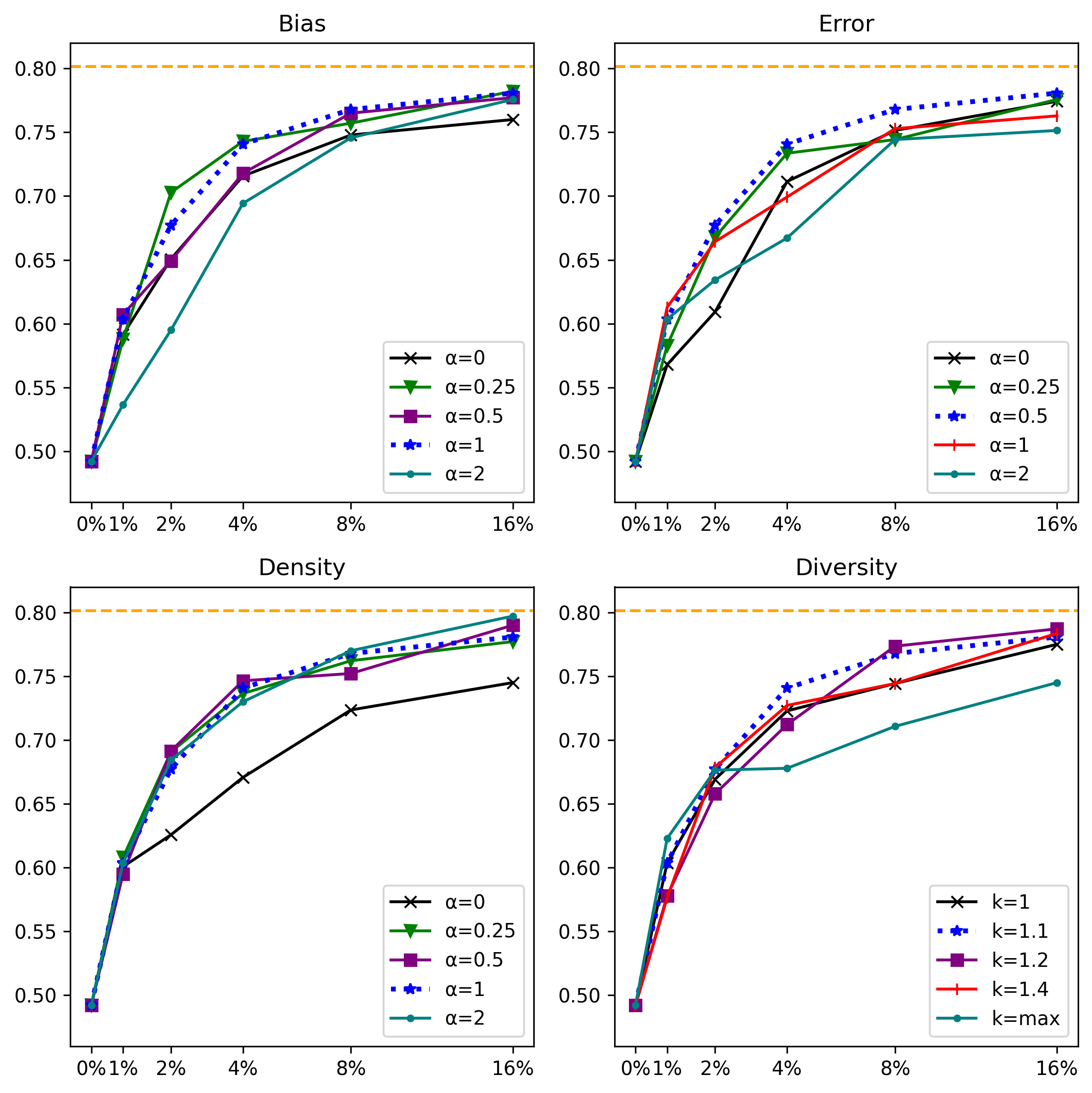}
    \caption{The parser accuracies across various query rounds on the \geode test set by employing the \amspnbest method. This method incorporates varying coefficients, denoted by $\alpha$, for each term. In addition to this, we also analyze the influence of the proportional rate, denoted by $k$, which represents the number of clusters in proportion to the budget size at each round.}
    \label{fig:sensitivity_result}
\end{figure}

Fig.~\ref{fig:sensitivity_result} shows the parser results on \geode test sets when we apply different coefficients or cluster sizes to the four independent acquisitions in \amspnbest. As we can see, tuning the parameters on an existing bilingual dataset does not necessarily bring optimal parser performance, indicating there is still potential in our approach if we can find suitable hyperparameter tuning methods. Another finding is that the \amspnbest is robust to the hyperparameters changes. Although changing the weights or cluster sizes for each term could influence the parser performances, the parser accuracies do not drop significantly. In addition, we have found that the Semantic Density and Semantic Diversity are more critical to \amspnbest as there are more fluctuations in the parser accuracies when we adjust the parameters of Semantic Density and Semantic Diversity.

\subsection{Parser Accuracies on English Test Sets}
\label{app:english_res}
\begin{figure}[ht]
    \centering
    \includegraphics[width=0.9\textwidth]{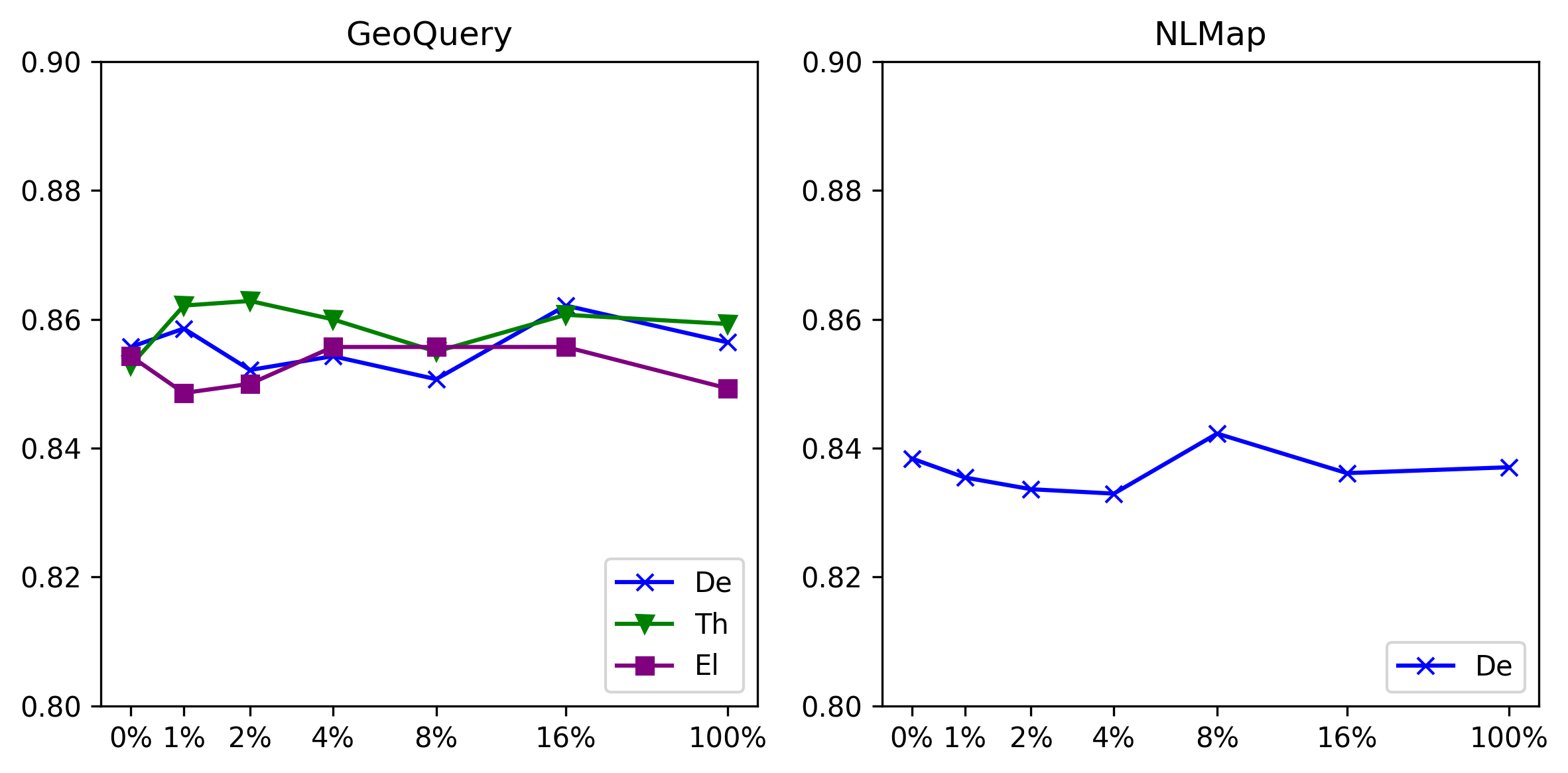}
    \caption{The accuracies on the English test sets after training the parsers on the training sets of \geode, \geoth, \geoel and \nlmapde acquired by \amspnbest at different query rounds.}
    \label{fig:eng_result}
\end{figure}

Fig.~\ref{fig:eng_result} shows the parser accuracies on the English test sets in different dataset settings. As we can see, the behaviours of the acquisition, \amspnbest, on the target languages do not influence the performance of parsers on the source languages.

\subsection{Ablation Study of Single Terms}
\label{app:only_term}
\begin{figure}[ht]
    \centering
    \includegraphics[width=0.9\textwidth]{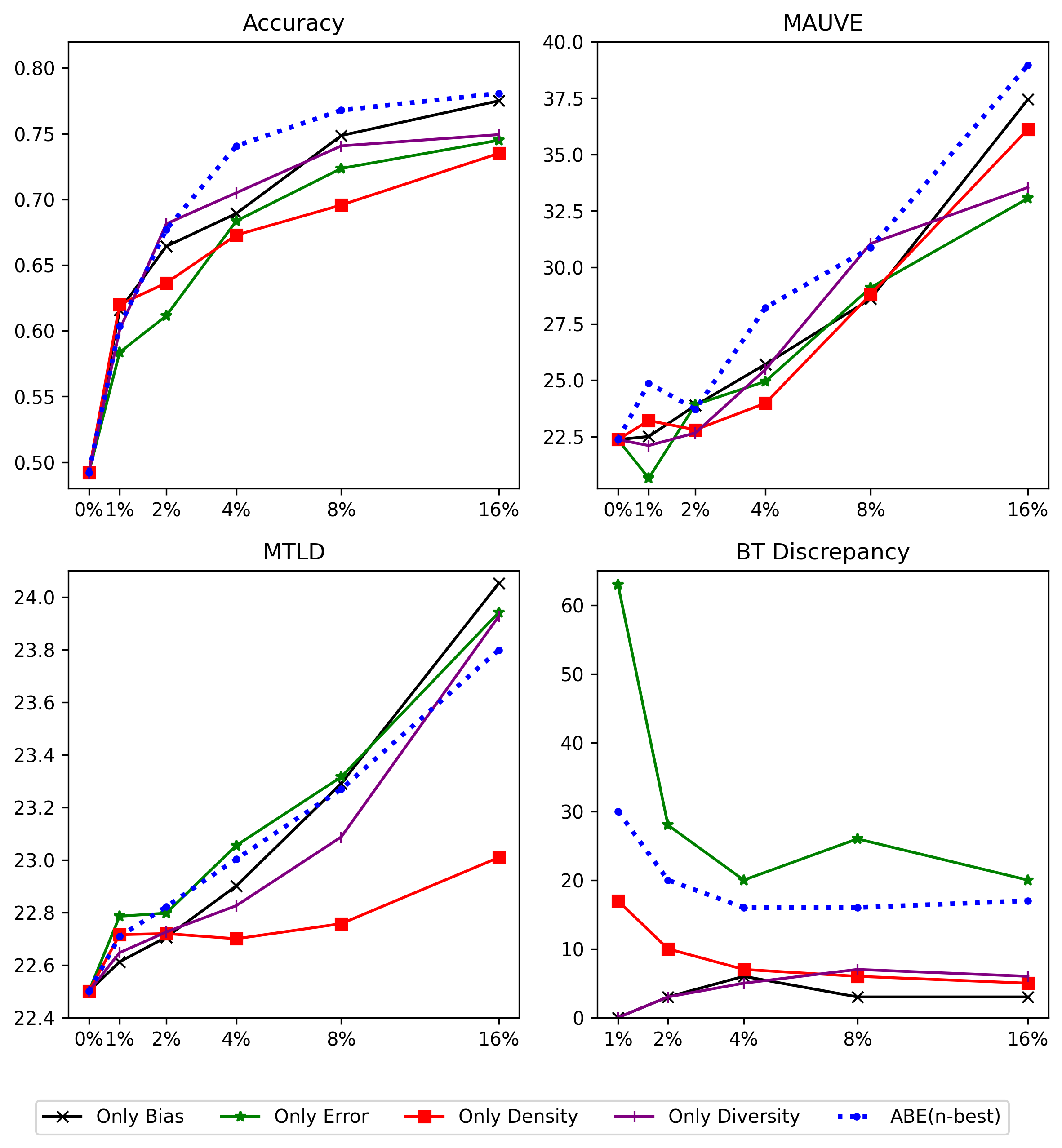}
    \caption{The parser accuracies at different query rounds using each single term in \amspnbest. \vspace{-4mm}}
    \label{fig:only_al_result}
        \vspace{-2mm}
\end{figure}

\subsection{Ablation Study of Factorization}
\label{app:abl_fact}
\begin{figure}[ht]
    \centering
    \includegraphics[width=0.9\textwidth]{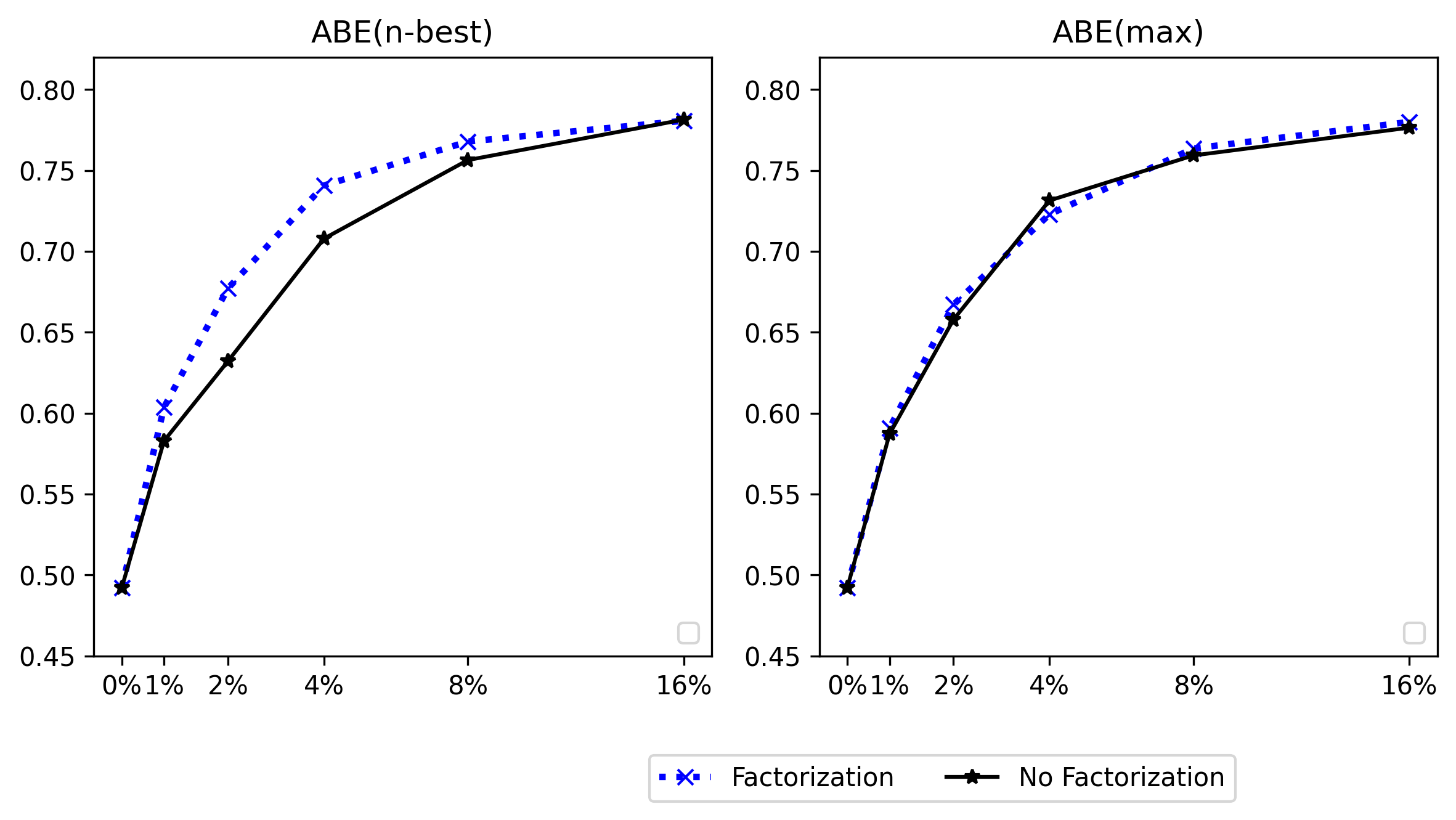}
    \caption{The parser accuracies using \amspnbest (Left) and \amspmax (Right) with or without factorization. \vspace{-4mm}}
    \label{fig:factorization_al_result}
        \vspace{-2mm}
\end{figure}

As in Fig.~\ref{fig:factorization_al_result}, at several query rounds, the parser accuracy can be 3\% - 6\% higher than that using no factorization in \textit{N-best Sequence Entropy}. But factorization does not help \amspmax improve the parser performance at all.

\subsection{Ablation Study of Error Term}
\label{app:error_term}
\begin{figure}[ht]
    \centering
    \includegraphics[width=0.9\textwidth]{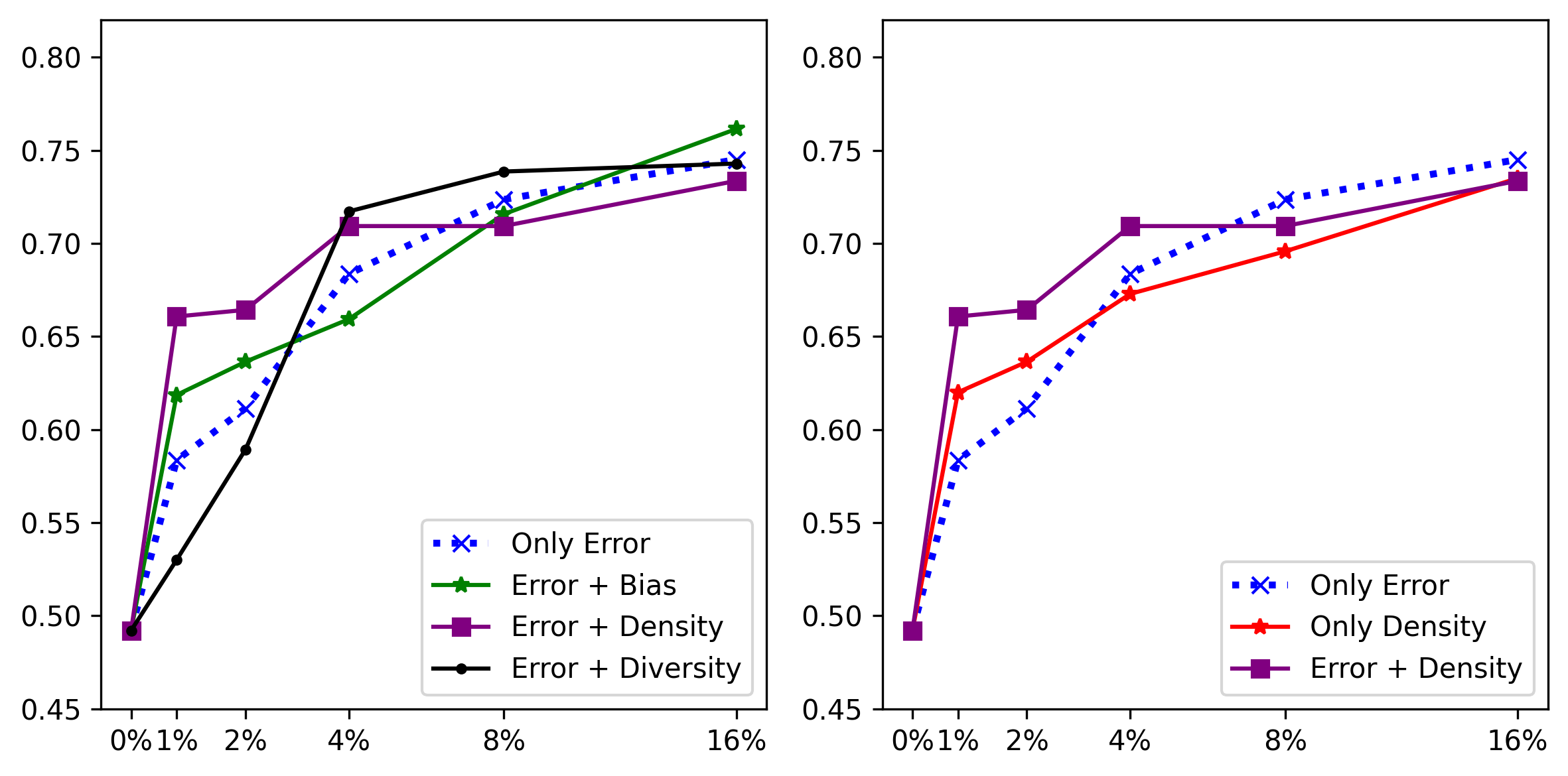}
    \caption{Combining Translation Error with different acquisition terms in \amspnbest, we depict the parser accuracies using different acquisition combinations at each query round. \vspace{-4mm}}
    \label{fig:error_result}
        \vspace{-2mm}
\end{figure}

As in Fig.~\ref{fig:error_result}, we combine Translation Error with different acquisition terms in \amspnbest. Combing Translation Error and Translation Error achieves the best result in the low sampling regions. The accuracy is even 4\% higher than the aggregated acquisition, \amspnbest, when the sampling rate is 1\%, suggesting that resolving translation error issues for semantically representative utterances benefits the parser more than resolving issues for tail utterances.

\subsection{BT Discrepancy Pattern}
\begin{figure}[ht]
    \centering
    \includegraphics[width=\textwidth]{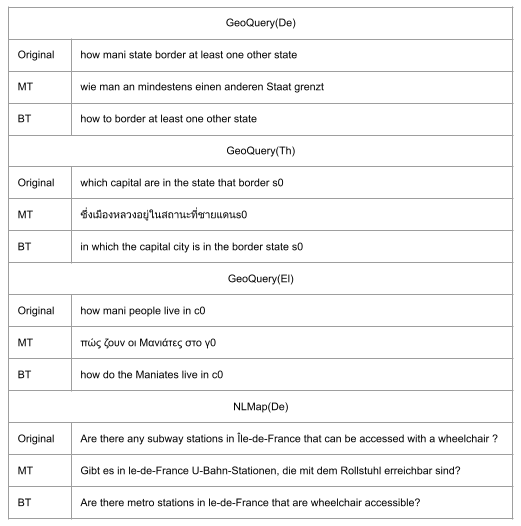}
    \caption{The original utterances and their corresponding machine translations and back-translations from \geode, \geoel, \geoth and \nlmapde.}
  \label{tab:back_examples}
\end{figure}

We observe that the BT discrepancy patterns vary as in Fig.~\ref{tab:back_examples}. For instance, the semantics of the BTs for Thai in \geo are altered dramatically due to the incorrect reorder of the words. Within \nlmap, the meanings of some German locations are inconsistent after the BT process. 

\subsection{T-SNE of Ablation Results}
\label{app:rep_picture}
\begin{figure*}
     \centering
     \begin{subfigure}[b]{0.495\linewidth}
         \centering
         \includegraphics[width=\linewidth]{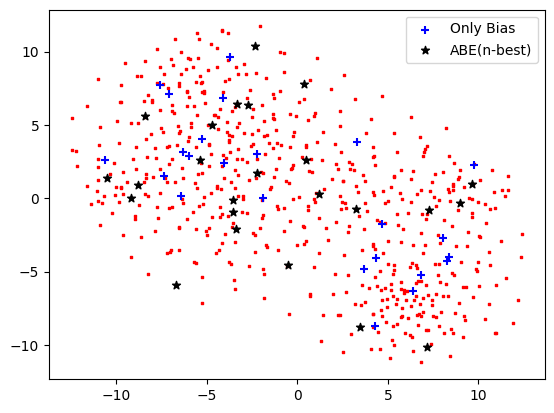}
     \end{subfigure}
     \hfill
     \begin{subfigure}[b]{0.495\linewidth}
         \centering
         \includegraphics[width=\linewidth]{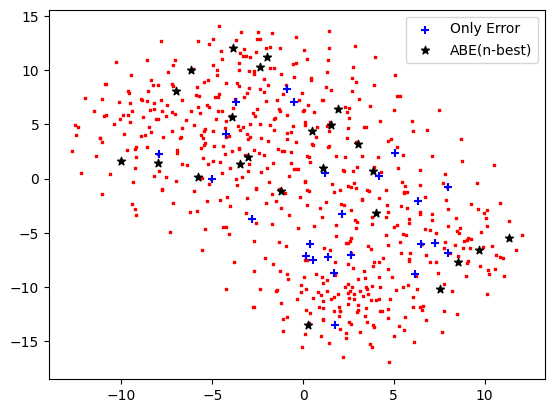}
     \end{subfigure}
     \hfill
          \begin{subfigure}[b]{0.495\linewidth}
         \centering
         \includegraphics[width=\linewidth]{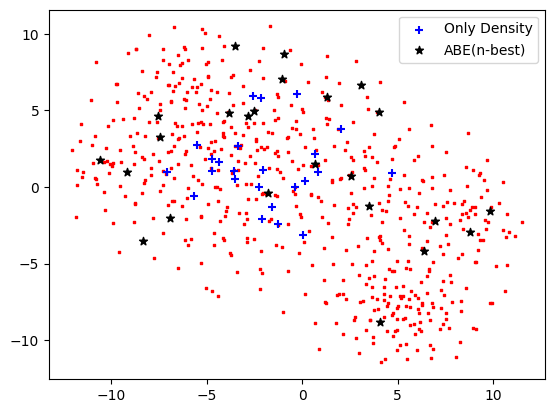}
     \end{subfigure}
     \hfill
     \begin{subfigure}[b]{0.495\linewidth}
         \centering
         \includegraphics[width=\linewidth]{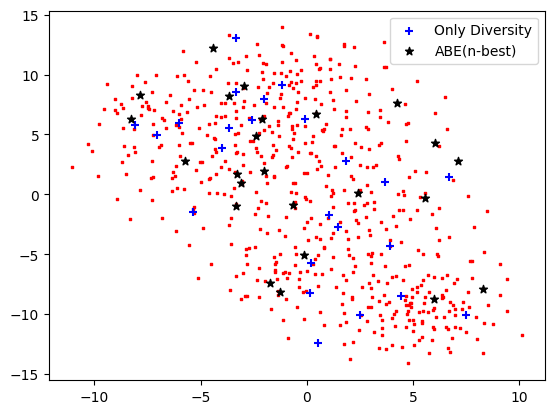}
     \end{subfigure}
         \hfill
     \begin{subfigure}[b]{0.495\linewidth}
         \centering
         \includegraphics[width=\linewidth]{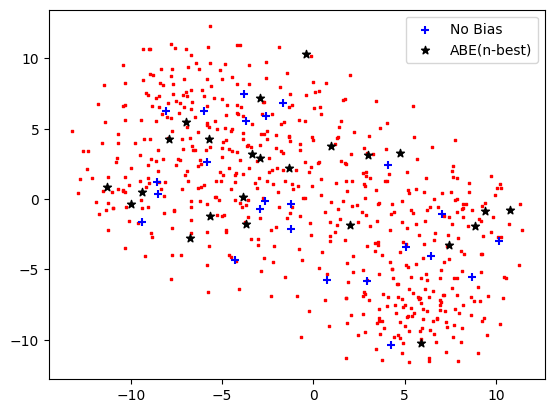}
     \end{subfigure}
     \hfill
     \begin{subfigure}[b]{0.495\linewidth}
         \centering
         \includegraphics[width=\linewidth]{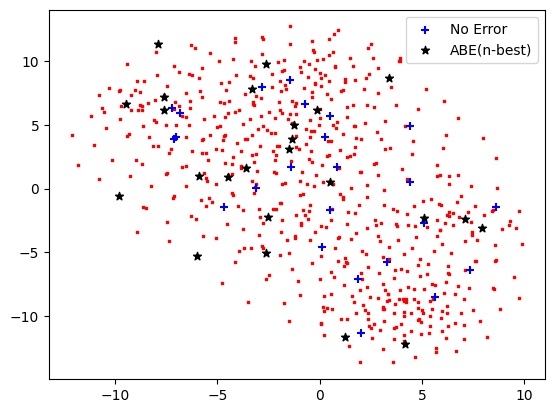}
     \end{subfigure}
     \hfill
          \begin{subfigure}[b]{0.495\linewidth}
         \centering
         \includegraphics[width=\linewidth]{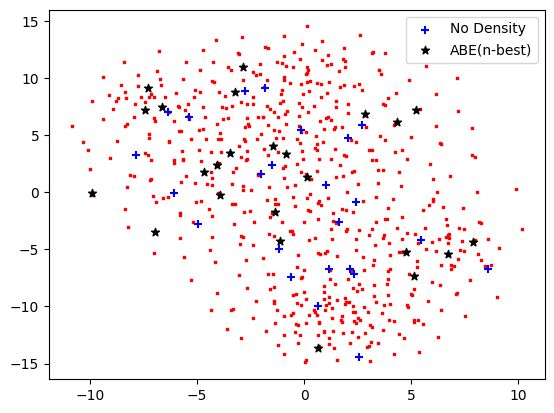}
     \end{subfigure}
     \hfill
     \begin{subfigure}[b]{0.495\linewidth}
         \centering
         \includegraphics[width=\linewidth]{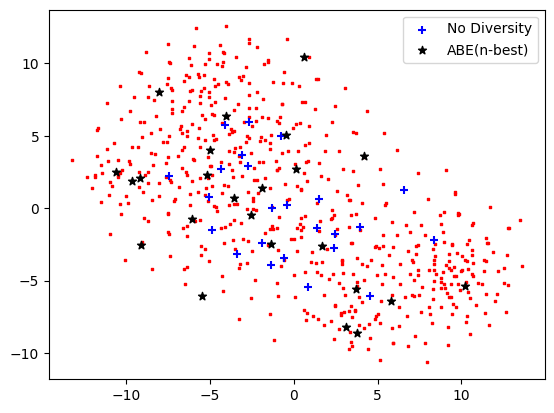}
     \end{subfigure}
     \hfill
     \hfill
        \caption{The representations of the English utterances selected by \amspnbest (black star) and the ablation settings (blue cross). Red dots are the representations of all the English utterances in the training set. }
        \label{fig:rep_picture}
\end{figure*}

Fig.~\ref{fig:rep_picture} shows the T-SNE plot of the representations of sampled utterances among all the utterances in the training set using \amspnbest and its various ablation settings. We encode the utterances with the pre-trained language model in the encoder of \bertlstm. We can see if we only use Semantic Density alone, the utterances are more likely to be in the dense region while not semantically diversified. On the contrary, the Translation Error tends to select tail utterances in the sparse semantic regions while they also lack semantic diversity. Both terms independently result in inferior parser performances. The Translation Bias and Translation Diversity collect utterances from diverse areas, thus providing better parser accuracies as in Fig.~\ref{app:only_term}. Removing Semantic Diversity from \amspnbest drops the parser performance most. As we observe, after removing Semantic Diversity, the utterances become more semantically similar compared to the utterances selected by \amspnbest. Overall, the T-SNE plot can be supplementary proof to our claim that we should retain the representativeness and diversity of the utterances to guarantee the parser performance.  

\onecolumn
\subsection{Evidence Lower Bound (ELBO)}
The maximum likelihood estimation objective of our parser is:

\begin{align}
\begin{split}
    & \argmax_{\theta}(\iiint_{\vx_s \in \mathcal{X}_{s}, \vx_t \in \mathcal{X}_{t}, \vy \in \mathcal{Y}} P_{\theta}(\vy, \vx_t, \vx_s)\,d\vx_s \,d\vx_t \,d\vy)
\end{split}
\end{align}

\noindent where $\vx_t$ is latent for most source utterance $\vx_s$. We assume $P_e(\vx_{t}|\vx_s)$ is a variational distribution.

\begin{align}
\begin{split}
    \log P_{\theta}(\vy, \vx_s)  & = \mathbb{E}_{P_e(\vx_{t}|\vx_s)}[ \log P_{\theta}(\vy, \vx_s)] \\
    & = \mathbb{E}_{P_e(\vx_{t}|\vx_s)}[ \log (\frac{P_{\theta}(\vx_t, \vy, \vx_s)}{P_{\theta}(\vx_t|\vy, \vx_s)})] \\
    \shortintertext{If we assume a conditional independence:}
    & \equiv \mathbb{E}_{P_e(\vx_{t}|\vx_s)}[ \log (\frac{P_{\theta}(\vx_t, \vy, \vx_s)}{P_{\theta}(\vx_t| \vx_s)})] \\
    & = \mathbb{E}_{P_e(\vx_{t}|\vx_s)}[ \log (\frac{P_{\theta}(\vx_t, \vy, \vx_s)}{P_e(\vx_t| \vx_s)} \frac{P_e(\vx_t| \vx_s)}{P_{\theta}(\vx_t| \vx_s)})] \\
    & = \mathbb{E}_{P_e(\vx_{t}|\vx_s)}[ \log (\frac{P_{\theta}(\vx_t, \vy, \vx_s)}{P_e(\vx_t| \vx_s)}] + \mathbb{E}_{P_e(\vx_{t}|\vx_s)}[\log \frac{P_e(\vx_t| \vx_s)}{P_e(\vx_t| \vx_s)})] \\
    & = \mathbb{E}_{P_e(\vx_{t}|\vx_s)}[ \log (\frac{P_{\theta}(\vx_t, \vy, \vx_s)}{P_e(\vx_t| \vx_s)})] + \KL(P_e || P_{\theta})
\end{split}
\end{align}

\noindent where $\mathbb{E}$ denotes the expectation function over a specified distribution and $\KL$ denotes the Kullback–Leibler divergence between two distributions. Therefore the ELBO of $\log P_{\theta}(\vy, \vx_s)$ is $\mathbb{E}_{P_e(\vx_{t}|\vx_s)}[ \log (\frac{P_{\theta}(\vx_t, \vy, \vx_s)}{P_e(\vx_t| \vx_s)})]$.

\begin{align}
\begin{split}
    ELBO(P_{\theta}(\vy, \vx_s)) & = \mathbb{E}_{P_e(\vx_{t}|\vx_s)}[ \log (\frac{P_{\theta}(\vx_t, \vy, \vx_s)}{P_e(\vx_t| \vx_s)})] \\
    & = \mathbb{E}_{P_e(\vx_{t}|\vx_s)}[\log P_{\theta}(\vx_t, \vy, \vx_s) - \log P_e(\vx_t|\vx_s)] \\
    & = \mathbb{E}_{P_e(\vx_{t}|\vx_s)}[\log P_{\theta}(\vy| \vx_t)] - \KL(P_e ||  P_{\theta}) + \mathbb{E}_{P_e(\vx_{t}|\vx_s)}[\log P_{\theta}(\vx_s)] \\
    & = \mathbb{E}_{P_e(\vx_{t}|\vx_s)}[\log P_{\theta}(\vy| \vx_t)] - \KL(P_e ||  P_{\theta}) + \log P_{\theta}(\vx_s)
\end{split}
\end{align}

\end{document}